\documentclass{article}
\usepackage{ijcai25}

\usepackage{times}
\usepackage{soul}
\usepackage{url}
\usepackage[hidelinks]{hyperref}
\usepackage[utf8]{inputenc}
\usepackage[small]{caption}
\usepackage{graphicx}
\usepackage{amsmath}
\usepackage{amsthm}
\usepackage{booktabs}
\usepackage{algorithm}
\usepackage{algorithmic}
\usepackage[switch]{lineno}
\usepackage{booktabs}  
\usepackage{soul}

\usepackage{algorithm}
\usepackage{algorithmic}
\usepackage{multirow}
\usepackage{color}
\usepackage{booktabs}
\usepackage{amssymb}
\usepackage{amsmath}
\usepackage{float}
\usepackage{subfig}

\usepackage{newfloat}
\usepackage{listings}
\usepackage{colortbl}
\definecolor{graycolor}{rgb}{0.95,0.95,0.95}
\title{Boosting Zero-shot Stereo Matching using Large-scale \\Mixed Images Sources  in the Real World}

\author{
Yuran Wang\thanks{These authors contributed equally.}\and
Yingping Liang$^{*}$\And
Ying Fu\thanks{Corresponding author: fuying@bit.edu.cn}\\
\affiliations
Beijing Institute of Technology
\emails
\{wangyuran, liangyingping, fuying\}@bit.edu.cn
}

\begin{document}

\maketitle

\renewcommand{\thefootnote}{\fnsymbol{footnote}} 

\begin{abstract}
Stereo matching methods rely on dense pixel-wise ground truth labels, which are laborious to obtain, especially for real-world datasets. The scarcity of labeled data and domain gaps between synthetic and real-world images also pose notable challenges. In this paper, we propose a novel framework, \textbf{BooSTer}, that leverages both vision foundation models and large-scale mixed image sources, including synthetic, real, and single-view images. First, to fully unleash the potential of large-scale single-view images, we design a data generation strategy combining monocular depth estimation and diffusion models to generate dense stereo matching data from single-view images. Second, to tackle sparse labels in real-world datasets, we transfer knowledge from monocular depth estimation models, using pseudo-mono depth labels and a dynamic scale- and shift-invariant loss for additional supervision. Furthermore, we incorporate vision foundation model as an encoder to extract robust and transferable features, boosting accuracy and generalization. Extensive experiments on benchmark datasets demonstrate the effectiveness of our approach, achieving significant improvements in accuracy over existing methods, particularly in scenarios with limited labeled data and domain shifts.
\end{abstract} 

\section{Introduction}

Stereo matching, the task of estimating disparity from two input images, plays an important role in computer vision, powering applications in fields such as robotics \cite{zhang2015building,zhang2024atlantis}, autonomous driving \cite{orb2017slam2}, and augmented reality \cite{yang2019security}. Recent advances in deep learning have led to the development of learning-based methods \cite{chang2018pyramid,shen2021cfnet,xu2023iterative} that achieve state-of-the-art performance. However, the generalization of these methods is highly dependent on the availability of high-quality and diverse training data. Existing datasets can be broadly categorized into synthetic and real-world datasets.

\begin{figure}[t]
    \centering
    \includegraphics[width=0.95 \linewidth]{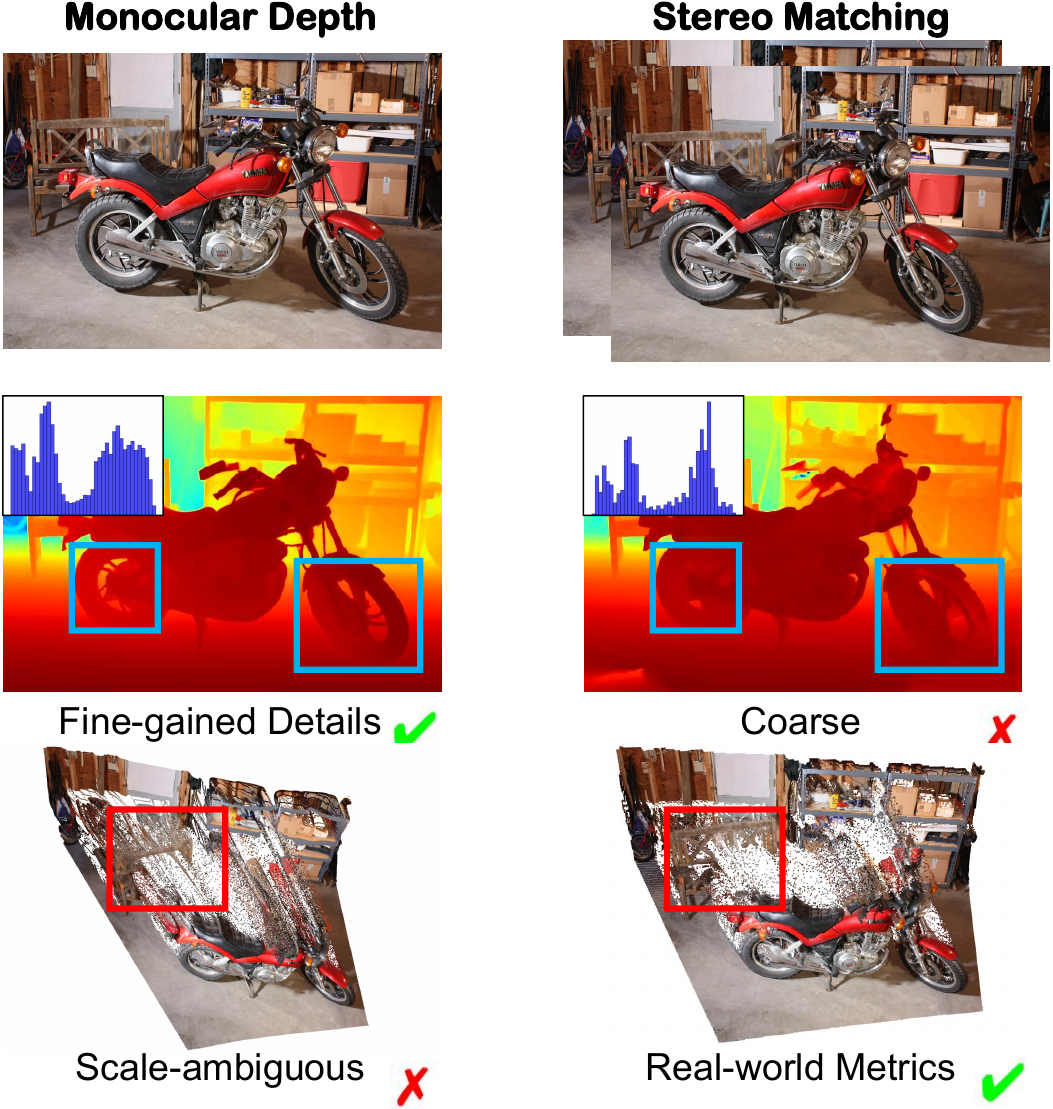}
    \caption{Comparison between Monocular Depth and Stereo Matching. Monocular Depth approach provides fine-grained details but suffers from scale ambiguity, whereas Stereo Matching delivers real-world metrics but with coarser results. We carefully design data generation and dynamic scale- and shift- invariant loss to perform knowledge transfer from monocular to stereo.}
    \label{teaser}
\end{figure}

Synthetic datasets, generated from blending engines \cite{mayer2016large}, mainly cover indoor environments. Models trained on these datasets perform well in indoor environments but show reduced effectiveness in outdoor scenarios due to greater visual and contextual variability. In contrast, real-world datasets \cite{geiger2012we,menze2015object} typically rely on LiDAR sensors for the ground truth disparity label. However, LiDAR-based ground truth is both costly and time-consuming to acquire and often suffers from issues such as misalignment and incompleteness. As a result, real-world stereo datasets often have sparse and noisy ground truth, which provides insufficient supervisory signals during training, limiting the generalization and accuracy of stereo matching models.

To address these challenges, several approaches have focused on refining network architectures \cite{xu2023iterative,lipson2021raft,zhangyk2025,zhang2021learning} or generating additional datasets \cite{liang2023mpi,guo2024lidar,zhang2023learning,wang2024mono2stereo}. Although these strategies lead to some performance improvements, they still do not resolve the domain gaps between synthetic and real-world data or address the sparsity of labels in real-world datasets well. Nowadays, the rise of large-scale vision foundation models (VFMs) has shown remarkable success and great generalization in a variety of tasks, including monocular depth estimation with models like DepthAnything \cite{depthanything} and segmentation with DINOv2 \cite{oquab2023dinov2}. Despite their success in numerous domains, the potential of VFMs for stereo matching remains underexplored. This is mainly because VFMs are typically designed for monocular tasks and do not explicitly incorporate the geometric constraints essential for stereo matching.

In this paper, we propose \textbf{BooSTer}, a novel framework that combines vision foundation models and large-scale mixed image sources, including synthetic, real-world, and single-view datasets, to \textbf{Boo}sting Zero-shot \textbf{STer}eo Matching performance. One of the key challenges in stereo matching is the scarcity of labeled data, particularly in real-world datasets. To address this, we introduce a monocular-depth-guided stereo data generation pipeline, which enriches training data from single-view images and allows the model to better generalize across diverse domains. Additionally, we tackle the issue of sparse and incomplete ground truth in real-world stereo datasets by using a monocular depth network to generate pseudo-labels and applying a dynamic scale- and shift-invariant loss to ensure robust learning of relative depth information. In addition, to bridge the domain gap between synthetic and real-world data, we integrate vision foundation model as an encoder to improve feature representations and boost the model’s generalization capability. Experimental results demonstrate the effectiveness of our method, showcasing its competitiveness in zero-shot stereo matching.

In summary, our main contributions can be summarized as follows:
\begin{itemize}
\item we present a novel framework that transfers knowledge from vision foundation models to stereo matching, boosting both generalization and performance.
\item we design a stereo data training framework guided by monocular depth estimation, enabling the creation of large-scale real-world stereo data and introducing a dynamic scale- and shift-invariant loss, to enhance stereo matching quality.
\item we employ a hybrid encoder combining a large-scale vision foundation model (DINOv2) and a traditional convolutional feature extractor to obtain global context and local details crucial for stereo matching.
\end{itemize}

\section{Related Work}

In this section, we review the most relevant studies on stereo matching, stereo dataset generation, and vision foundation models.

\subsection{Stereo Matching Method}

Learning-based stereo matching methods have been replaced traditional optimization methods with CNN networks. GCNet~\cite{kendall2017end} first uses 3D convolutional encoder-decoder architecture to regularize a 4D volume. Following the success of GCNet, PSMNet~\cite{chang2018pyramid}, GwcNet~\cite{guo2019group} and GANet~\cite{zhang2019ga} gradually increase the precision of the network. Besides, cascade methods like CFNet~\cite{shen2021cfnet} is proposed to improve efficiency. RAFT-Stereo~\cite{lipson2021raft} proposes to recurrently update disparity map using local cost values retrieved from the all-pairs correlations. IGEV~\cite{xu2023iterative} constructs a new module to encode non-local geometry and context information. StereoBase~\cite{guo2023openstereo} serves as a strong baseline model in deep stereo-matching by combining all existing architectures. While these methods have increasingly complex network structures, none has focused on improving the encoder architecture, which is crucial for stereo matching performance.

\begin{figure*}[ht!]
    \centering
    \includegraphics[width = 0.95\linewidth]{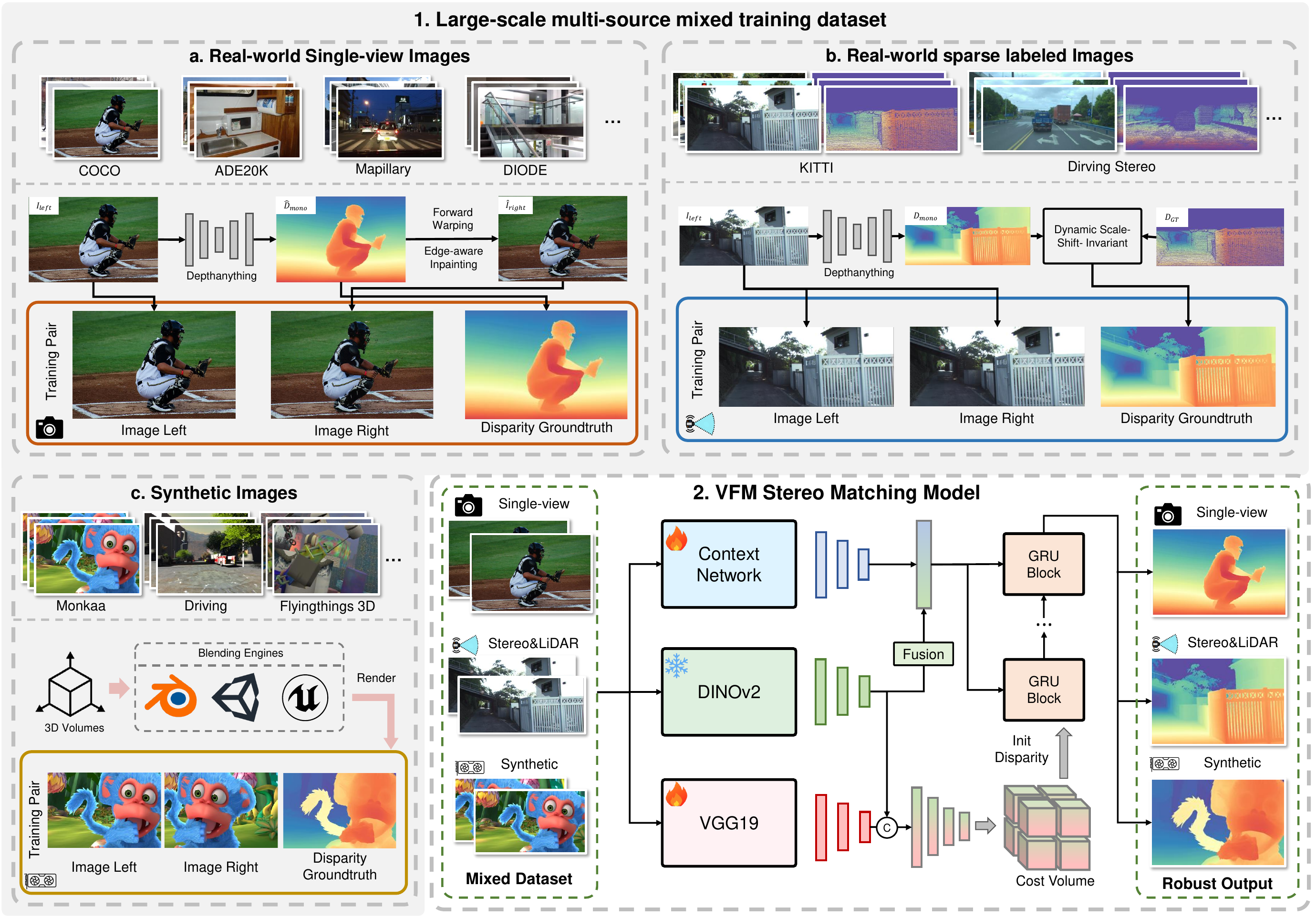}
    \caption{The overall architecture of our proposed method consists of two main parts. 1. \textbf{Large-scale multi-source mixed training dataset.} We mix stereo data from various scenarios as a large-scale pre-training dataset. 2. \textbf{VFM Stereo Matching Model.} We propose a hybrid encoder structure embedded in VFM to boost the generalization and performance of the algorithm by transferring existing knowledge.}
    \label{fig:framework}
\end{figure*}
\subsection{Stereo Datasets}
Stereo datasets can be categorized into synthetic and real-world datasets. Real-world datasets, such as KITTI12~\cite{geiger2012we} and KITTI15~\cite{menze2015object}, provide 200 pairs of outdoor images with sparse disparity labels derived from LiDAR points, while the ETH 3D~\cite{schops2017multi} dataset focuses on indoor scenes captured with depth cameras. Recent large stereo datasets, including DIML~\cite{cho2021diml}, HRWSI~\cite{xian2020structure}, and IRS~\cite{wang2021irs}, offer stereo data, though their ground truth (GT) labels are obtained via stereo matching, which limits the performance of deep stereo networks. Synthetic datasets, like SceneFlow~\cite{mayer2016large}, generate stereo images using computer graphics (CG) and provide ground truth disparity maps. However, CG methods struggle to model complex real-world scenes, resulting in domain gaps. Attempts like Mono-for-Stereo (MfS)~\cite{watson2020learning}, which generates stereo pairs from single-view images, often suffer from realism issues such as collisions, holes, and artifacts, negatively affecting network performance.


\subsection{Vision Foundation Models}

With advancements in computing power and the growth of large-scale datasets, a growing number of vision foundational models (VFMs)~\cite{depthanything,radford2021learning,ravi2024sam2,oquab2023dinov2} with strong generalization and high performance have emerged. For instance, in the area of monocular depth estimation, models like DepthAnything~\cite{depthanything} have demonstrated outstanding generalization performance. Moreover, numerous studies have introduced more general backbone networks through self-supervised pre-training. Among these, the DINO family~\cite{oquab2023dinov2,darcet2023vitneedreg} has explored self-supervised learning with visual transformers, achieving improvements across a range of downstream tasks. Despite the impressive generalization and zero-shot capabilities of these VFMs, how to effectively leverage them to enhance stereo matching performance remains an open challenge.
\section{Method}
In this section, we first introduce the motivation and formulation of BooSTer. Then, we present the details of our DFM Guided Training Framework and VFM integrated hybrid encoder architecture.
\subsection{Motivation and Formulation}
In the stereo matching task, the input consists of a pair of stereo images, $\mathbf{I}_l$ and $\mathbf{I}_r$. The goal is to generate a dense disparity map $\mathbf{D}_{pred}$. To achieve this, a stereo matching model $f_{\theta}$ parameterized by $\theta$ is commonly trained as
\begin{align}
\mathbf{D}_{pred} &= f_{\theta}(\mathbf{I}_l, \mathbf{I}_r).
\end{align}

To guide the model’s learning, a sparse supervision loss is defined based on the available sparse depth annotations
\begin{align}
\mathcal{L}_{sup} &= \|\mathbf{M}_{val} \odot (\mathbf{D}_{GT} - \mathbf{D}_{pred})\|_1,
\end{align}
where $\mathbf{D}_{GT}$ denotes the ground truth disparity map, and $\mathbf{M}$ represents the valid mask for the ground truth disparity $\mathbf{D}_{GT}$. $\mathbf{M}_{val}$ is a matrix of valid masks for disparity maps. In practice, the stereo matching model $f_{\theta}$ consists of two components, \emph{i.e.}, a feature extraction encoder $f_e$, and a disparity refinement module $f_d$. 

The performance and generalization of stereo matching algorithms are often constrained by the limited scale of real-world datasets and the encoder's challenge in learning robust features from synthetic data. While synthetic datasets provide large amounts of data, they fail to capture real-world complexities, such as lighting, texture, and noise, making models trained on them struggle to generalize effectively to real-world environments.

To address these limitations, we aim to improve both generalization and performance by expanding the training dataset and incorporating a hybrid encoder. We introduce a large-scale mixed dataset of real-world images alongside synthetic datasets and integrate vision foundation model into our encoder. This hybrid encoder leverages the pre-trained semantic and feature extraction capabilities of VFM, enabling more accurate and comprehensive feature extraction and enhancing the model’s generalization and performance.


\begin{figure}[t]
\centering
\subfloat[Warping Image]{\includegraphics[width=0.32 \linewidth]{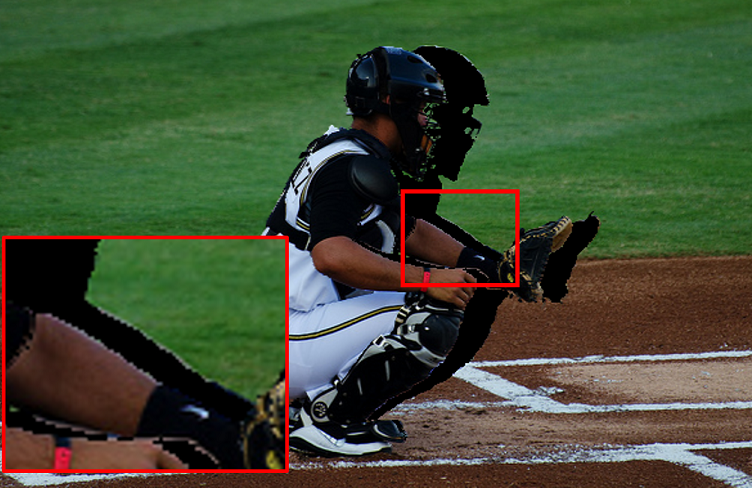}}
\hfill
\subfloat[Inpaint w/o EA]{\includegraphics[width=0.32 \linewidth]{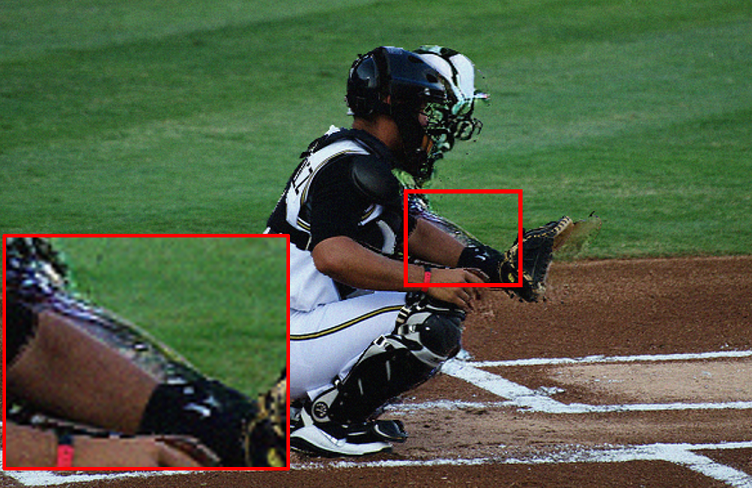}}
\hfill
\subfloat[Inpaint w/ EA]{\includegraphics[width=0.32 \linewidth]{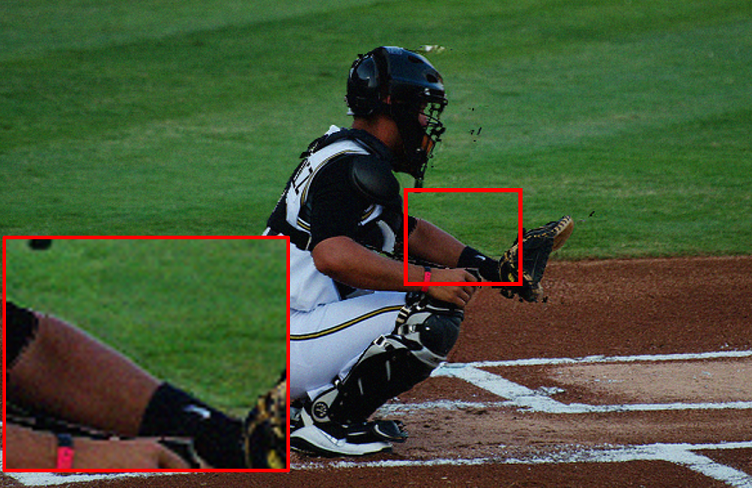}}
\caption{Comparison between naive inpainting module and our inpainting module. (a) warped right-view image with occulusion holes, (b) image inpainted from (a) using SD, and (c) right image warping with our inpainting module.}
\label{Inpainting}
\end{figure}

\subsection{Learning from Single-view Images}

A disparity map is defined as the per-pixel horizontal displacement between the corresponding locations of every pixel from the first view to the second. In this case, from the left image $\mathbf{I}_l$ to the right image $\mathbf{I}_r$. Described in mathematical form, the disparity map can be defined as
\begin{eqnarray}
\mathbf{D}(i) = x_l(i) - x_r(i')
\end{eqnarray}
where $\mathbf{D}(i)$ is the disparity value of pixel $i$, and $x_l(i)$ and $x_r(i')$ are the horizontal coordinates of pixel $i$ and $i'$ in the left and right view images, respectively.

Given any disparity map, we can warp every left view pixel $i$ to the right view pixel $i'$ using the disparity map. Therefore, we use the monocular depth estimation model to formulate the disparity map $\mathbf{D}_{mono}$. The disparity map from the monocular model is a relative disparity map ranging from $[0,1]$. We scale $\mathbf{D}_{mono}$ with a random factor $\alpha$ to convert it into the pixel-wise disparity map $\mathbf{D}_{mono}'$. The transformation can be formulated as
\begin{eqnarray}
    \mathbf{D}_{mono}' = \alpha \mathbf{D}_{mono},
\end{eqnarray}
where $\alpha \in [d_{min}, d_{max}]$. To boost the diversity of the generated data, we randomly sample the scaling factor $alpha$ from a uniform distribution $U(d_{min}, d_{max})$, where $d_{min}$ and $d_{max}$ are the minimum and maximum scaling factors, respectively.


\noindent\textbf{Edge-Aware Inpainting Module.} 
Random scaling and forward warping address the challenge of generating stereo image pairs. However, the generated right-view images often contain occlusion holes in regions that are invisible in the left-view images. To fill in these missing parts, we use Stable Diffusion (SD). However, as shown in Figure \ref{Inpainting}, simply using an inpainting model tends to blend the front and back contents. To address this issue, we present an Edge-Aware (EA) inpainting module to mitigate the blending problem. According to Figure \ref{Inpainting}, this blending issue arises from a lack of object edge information. Therefore, we first generate the object edge mask $\mathbf{M}_e \in [0,1]^{H \times W}$ from $\mathbf{D}_{mono}'$. We detect the horizontal object edges to create the edge mask using
\begin{eqnarray}
    \mathbf{M}_e(i) = \left\{
    \begin{array}{ll}
    1, & if \  \nabla_x(\mathbf{D}_{mono}'(i)) > \tau \\
    0, & otherwise
    \end{array}
    \right.,
\end{eqnarray}
where  $\nabla_x(\cdot)$ computes the horizontal gradient of the disparity map, and $\tau$ is the threshold for detecting object edges. We select a few background pixels from $\mathbf{M}_e$ and warp them with the foreground to preserve edge information, and then apply inpainting with the SD model to fill occultation holes. The EA module effectively reduces foreground-background fusion, enhancing the realism of the inpainted image.

\begin{figure}[t]
\centering
\subfloat[RGB Image]{\includegraphics[width=0.32 \linewidth]{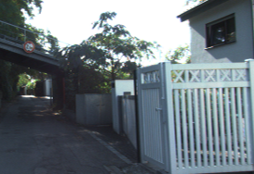}}
\hfill
\subfloat[GT Disparity]{\includegraphics[width=0.32 \linewidth]{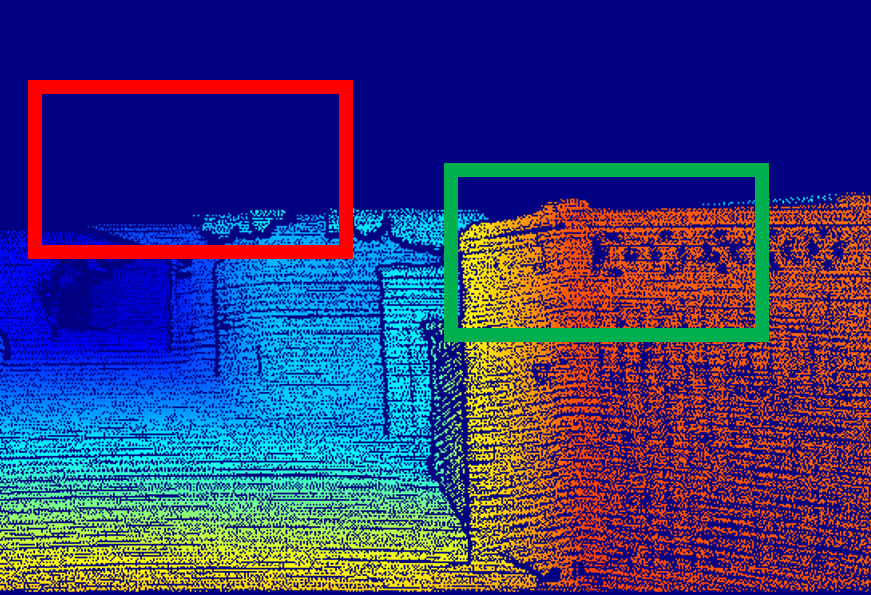}}
\hfill
\subfloat[Mono-Prediction]{\includegraphics[width=0.32 \linewidth]{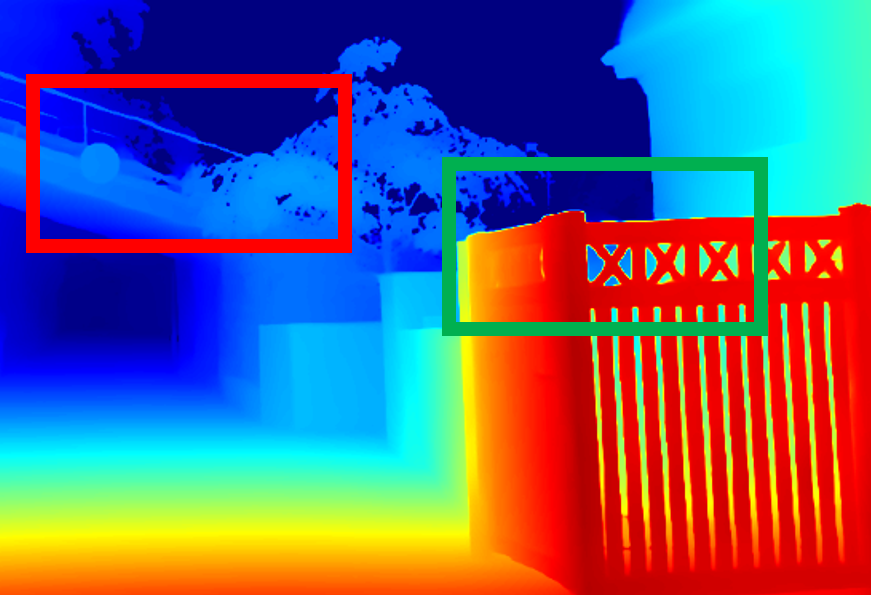}}
\caption{Comparison between disparity from KITTI and from monocular model. (a) RGB Image, (b) ground-truth from LiDAR, and (c) disparity from monocular estimation.}
\label{pseudo labels}
\end{figure}

\subsection{Learning from Sparse Images}

As shown in Figure \ref{pseudo labels}, pseudo-labels from monocular depth estimation provide accurate details and infer relative depth. However, these scaleless depth estimates are not suitable as direct supervision for stereo models. Therefore, we propose to transfer both monocular depth estimation and stereo matching model output into same scaleless domain to fix the domain gap. As a result, we introduce dynamic scale- and shift- invariant (DSSI) transformation to align the prediction to monocular pseudo depth based on a least-squares criterion
\begin{equation}
(a,b) =\arg\min_{a,b}\sum_{i=1}^M\left(a\mathbf{D}_{pred}(i)+b-\mathbf{D}_{mono}(i)\right)^2, 
\end{equation}
where $a$ is scale factor and $b$ is shift factor. Due to the susceptibility of the least squares method to outliers, which compromises the stability of supervision, we conducted outlier detection and removal on the scaled map then recalculated the $a$ and $b$. The specific mathematical expressions are
\begin{align}
    \mathbf{M}_l(i) =& \left\{
    \begin{array}{ll}
    1, & if \  (a\mathbf{D}_{pred}(i)+b-\mathbf{D}_{mono}(i))^2 < \tau_l \\
    0, & otherwise
    \end{array}
    \right.,\\
    \tau_l =& Q(a\mathbf{D}_{pred}+b-\mathbf{D}_{mono}),
\end{align}
where $\tau_l$ is the threshold of masking, and $Q(\cdot)$ is the Quintiles operation. After filtering, we recompute the scale $a'$ and shift $b'$ based on the remaining value and develop DSSI loss using mean-squared error (MSE), 
\begin{equation}
\begin{aligned}
\mathcal{L}_{DSSI} &= MSE(\hat{\mathbf{D}}_{pred}, \mathbf{D}_{mono}),\\
\hat{\mathbf{D}}_{pred} &= a'\mathbf{D}_{pred} + b'.
\end{aligned}
\end{equation}
Together with the sparse GT loss, the final loss function is formulated as 
\begin{eqnarray}
    \mathcal{L} = \mathcal{L}_{sparse} + \beta\mathcal{L}_{DSSI},
\end{eqnarray}
where $\beta$ is the DSSI factor of loss functions.

\begin{table*}[ht!]
\centering
\small
\begin{tabular}{@{}lcccccccccccc@{}}
\toprule
\multirow{2}{*}{Networks}  & \multicolumn{3}{c}{KITTI12} & \multicolumn{3}{c}{KITTI15} & \multicolumn{3}{c}{ETH 3D} & \multicolumn{3}{c}{Middlebury} \\
\cmidrule(lr){2-4} \cmidrule(lr){5-7} \cmidrule(lr){8-10} \cmidrule(lr){11-13} 
& D1 $\downarrow$ & EPE $\downarrow$ & \textgreater2px $\downarrow$ & D1 $\downarrow$ & EPE $\downarrow$ & \textgreater2px $\downarrow$ & D1 $\downarrow$ & EPE $\downarrow$ & \textgreater2px $\downarrow$ & D1 $\downarrow$ & EPE $\downarrow$ & \textgreater2px $\downarrow$\\
\midrule
PSMNet'18 & 30.51 & 4.68 & 42.15 & 32.14 & 5.98 & 44.26 & 8.79 & 8.26 & 10.20 & 22.54 & 7.93 & 33.53 \\
CFNet'21 & 13.63 & 2.27 & 20.07 & 12.08 & 2.89 & 19.04 & 3.07 & 2.48 & 4.03 & 18.41 & 6.60 & 23.91\\
GwcNet'19 & 23.04 & 2.76 & 33.71 & 25.19 & 3.57 & 36.96 & 6.31 & 2.59 & 7.94 & 22.16 & 7.28 & 29.87\\
COEX'21 & 12.07 & 1.79 & 20.23 & 11.00 & 2.48 & 19.98 & 3.71 & 1.74 & 5.37 & 18.19 & 6.82 & 25.16\\
FADNet++'21 & 11.31 & 1.77 & 18.02 & 13.23 & 2.97 & 20.90 & 11.73 & 8.36 & 14.26 & 15.23 & 4.87 & 24.08\\
CasStereo'20 & 11.85 & 1.82 & 18.84 & 12.06 & 2.69 & 20.26 & 3.83 & 1.48 & 5.42 & 20.61 & 7.58 & 27.39 \\
IGEV'23 & 4.81 & 0.95 & 7.84 & 5.10 & 1.20 & 8.90 & 1.39 & 0.45 & 1.88 & 8.44 & 2.10 & 11.77\\
Selective-IGEV'24 & 5.73 & 1.08 & 5.73 & 5.63 & 1.24 & 9.60 & 2.45 & 1.35 & 3.03 & 24.81 & 23.63 & 29.11 \\
StereoBase'24 & 4.98 & 1.02 & 8.09 & 5.47 & 1.20 & 9.47 & 1.28 & 0.28 & 1.76 & 8.31 & 1.71 & 11.95  \\
\rowcolor{graycolor}\textbf{Ours}& \textbf{3.04} & \textbf{0.76} & \textbf{5.19} & \textbf{3.22} & \textbf{0.89} & \textbf{6.63} & \textbf{0.70} & \textbf{0.24} & \textbf{1.12} & \textbf{7.50} & \textbf{1.66} & \textbf{11.72}\\
\bottomrule
\end{tabular}
\caption{Comparison with stereo matching methods. ``Ours" indicates pre-training on our mixed dataset. Models are validated on the KITTI12, KITTI15 and ETH3D training sets as cross-domain validation. By default, all supervised methods have trained using SceneFlow.}
\label{tab:finetune-test}
\end{table*}

\subsection{VFM Based Hybrid Feature Encoder}

With the rapid advancement of deep learning techniques, numerous network architectures have been proposed for stereo matching task, each incorporating distinct module designs. Most stereo matching networks can generally be categorized into a two-stage framework, a feature extraction encoder, and a disparity refinement module based on cost volume.

Here, we employ a hybrid feature extraction encoder by integrating DINOv2, a powerful pre-trained feature extraction model, with the traditional convolutional neural network (CNN) based encoder. DINOv2 captures richer and more robust high-dimensional feature representations, including both semantic and high-level information, while CNN focuses on extracting detailed pixel-wise features. This combination enables the model to process the image from both global and local perspectives, thereby enhancing the performance and effectiveness of the encoder.

Formally, given a pair of left and right images $\mathbf{I}_l, \mathbf{I}_r \in \mathbb{R}^{H \times W \times 3}$, we begin by employing the well-established VGG19~\cite{vgg} as the CNN feature extractor. This network generates multi-level pyramid features, producing feature maps at various scales: $f_c^{(i)} \in \mathbb{R}^{C_i \times \frac{H}{i} \times \frac{W}{i}}$, where $i \in \{4, 8, 16\}$ denotes the scaling factor. These multi-scale features offer a rich and layered representation of the input images. To further enhance the feature extraction process, we introduce DINOv2 to generate an additional high-dimensional feature layer, $f_c^{(32)} \in \mathbb{R}^{C_{32} \times \frac{H}{32} \times \frac{W}{32}}$, which captures high-level semantic and contextual information essential for stereo matching.

Finally, the disparity refinement network is designed following the methodology outlined in~\cite{guo2023openstereo}, which can boost the depth estimation by refining the initial disparity maps. The combination of advanced feature extraction and disparity refinement improves accuracy and generalization in stereo matching.


\section{Experiment}

In this section, we first introduce the datasets used for training and evaluation, as well as the details of our implementation. Then, detailed comparisons are conducted under zero-shot settings. Finally, ablation studies are performed to confirm the effectiveness of our proposed main components.
\subsection{Datasets and Evaluation Metrics}

\noindent\textbf{ETH3D}~\cite{schops2017multi} is a widely used dataset with grayscale stereo pairs from indoor and outdoor environments, featuring LiDAR ground truth. It includes 27 training and 20 test frames for low-resolution two-view stereo. We use ETH3D as validation set to evaluate our method.

\noindent\textbf{KITTI 2012 and KITTI 2015}~\cite{geiger2012we,menze2015object} are well-known benchmarks with 200 labeled training pairs and additional test pairs. Ground truth is provided by sparse LiDAR. KITTI15 includes dynamic scenes with semi-automatically generated ground truth. We evaluate the trained models on the training sets of both datasets, following prior work.

\noindent\textbf{Middlebury}~\cite{scharstein2014high} contains two sets of 15 stereo image pairs for training and testing, captured from indoor scenes at three resolutions (full, half, and quarter). For evaluations, only the training pairs at half resolution are used to assess cross-domain generalization.

\noindent\textbf{Mixed training datasets} consist of three component. The first is synthetic data, such as SceneFlow~\cite{mayer2016large}, which includes over 39,000 stereo frames at $960\times540$ resolution, but has domain gaps with real-world applications. To address this, we create the Diffusion-based Mono for Stereo (\textbf{DiffMFS}) dataset by combining single-view datasets like COCO 2017~\cite{lin2014microsoft}, Mapillary Vistas~\cite{neuhold2017mapillary}, ADE20K~\cite{zhou2017scene}, Depth in the Wild~\cite{chen2016single}, and DIODE~\cite{vasiljevic2019diode}, resulting in 597,727 single-view images. Our method then synthesizes stereo data for training. For real-world datasets, we use Driving Stereo~\cite{yang2019drivingstereo} and apply our dynamic scale- and shift-invariant loss to handle missing and sparse LiDAR ground truth. In each training epoch, we randomly sample images from the three datasets with a sampling frequency of 5:6:1.

\begin{figure*}[]
\centering
\small
\setlength\tabcolsep{1pt}
\begin{tabular}{cccc}
\includegraphics[width=0.245\textwidth]{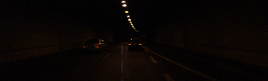} &  
\includegraphics[width=0.245\textwidth]{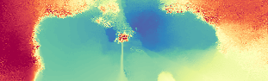} &  
\includegraphics[width=0.245\textwidth]{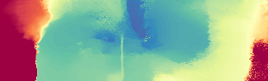} & 
\includegraphics[width=0.245\textwidth]{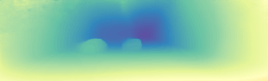} \\
\includegraphics[width=0.245\textwidth]{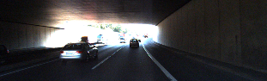} & 
\includegraphics[width=0.245\textwidth]{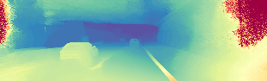} &  
\includegraphics[width=0.245\textwidth]{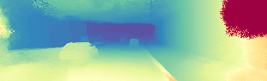} &  
\includegraphics[width=0.245\textwidth]{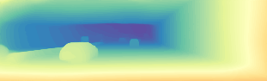} \\
\includegraphics[width=0.245\textwidth]{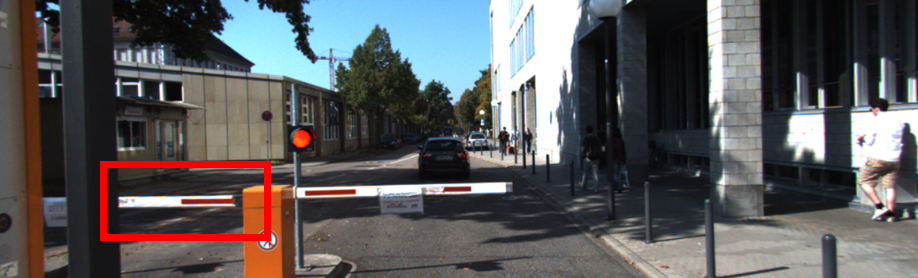} &  
\includegraphics[width=0.245\textwidth]{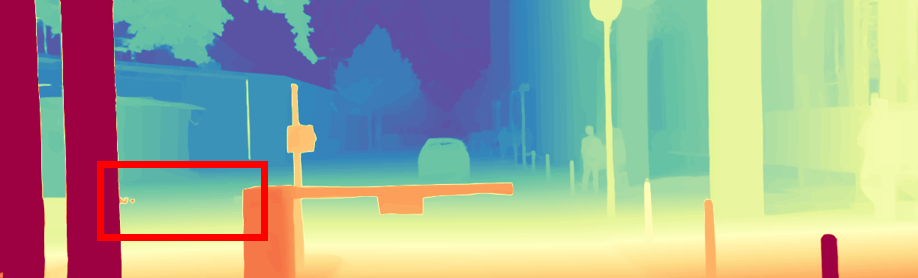} &  
\includegraphics[width=0.245\textwidth]{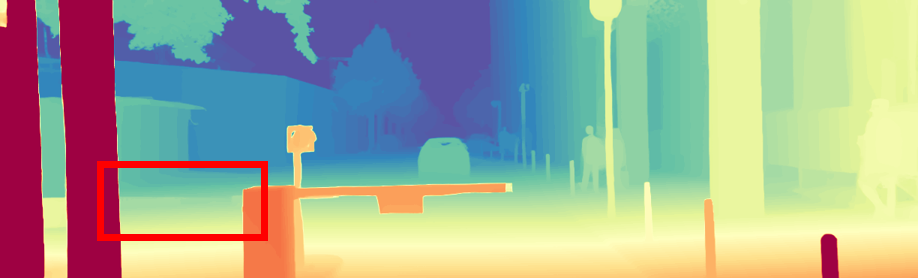} &  
\includegraphics[width=0.245\textwidth]{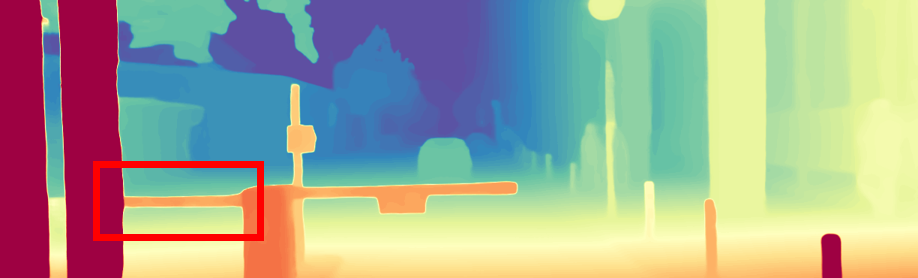} \\
\includegraphics[width=0.245\textwidth]{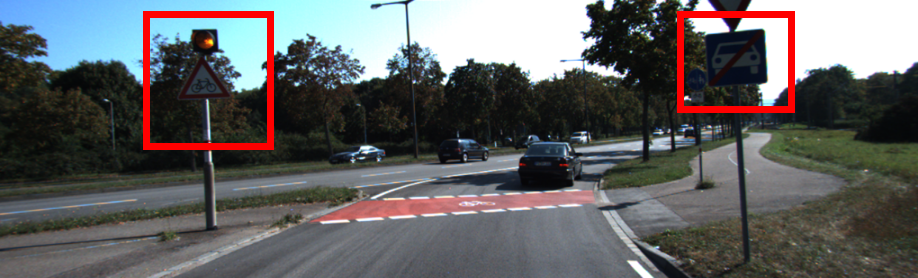} &  
\includegraphics[width=0.245\textwidth]{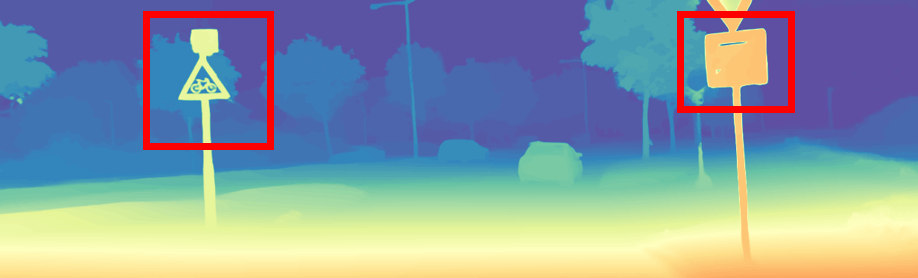} &  
\includegraphics[width=0.245\textwidth]{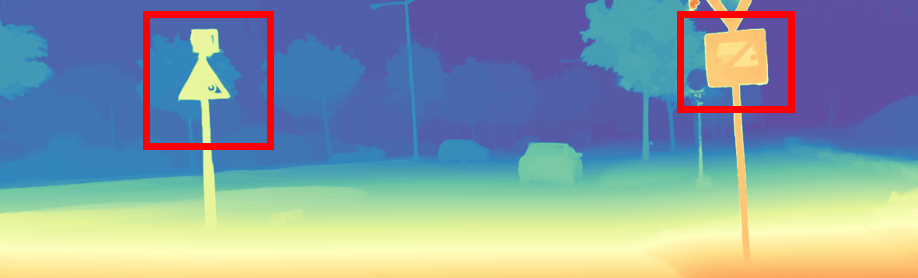} &  
\includegraphics[width=0.245\textwidth]{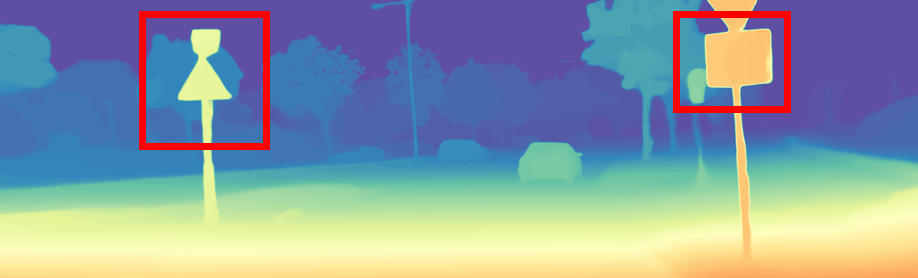} \\
\includegraphics[width=0.245\textwidth]{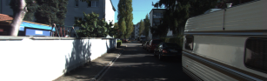} &  
\includegraphics[width=0.245\textwidth]{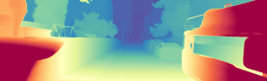} &  
\includegraphics[width=0.245\textwidth]{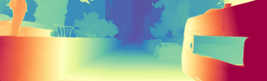} &  
\includegraphics[width=0.245\textwidth]{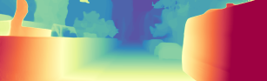} \\
\includegraphics[width=0.245\textwidth]{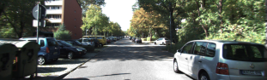} &  
\includegraphics[width=0.245\textwidth]{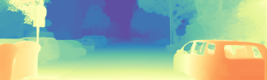} &  
\includegraphics[width=0.245\textwidth]{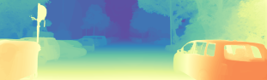} & 
\includegraphics[width=0.245\textwidth]{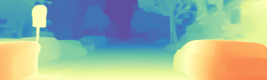} \\
Left Image & IGEV & StereoBase & Ours
\end{tabular}
\caption{Qualitative results of IGEV and StereoBase trained with Sceneflow and our model trained on our mixed dataset. By default, IGEV and StereoBase are trained using SceneFlow.}
\label{fig:vis-dataset}
\end{figure*}

\noindent\textbf{Evaluation Metrics.} We report evaluation results on the average End-Point Error (EPE), D1, and $>$2px metric, which indicates the percentage of stereo disparity outliers. For all metrics, smaller values indicate better model performance.
\subsection{Implementation Details}

We follow the setup of recent work~\cite{guo2023openstereo} and train our model on 4 NVIDIA 4090 GPUs. For the DFM used in stereo image generation, we adopt Depth Anything V2~\cite{depthanything} due to its great generalization. For VFM in the hybrid encoder, we use DINOv2~\cite{oquab2023dinov2} as its powerful image understanding capabilities and robust features. In the training stage, we utilize our mixed training dataset. We adopt AdamW with a weight decay of 0.05 as the optimizer and clip gradients if their norms exceed 0.1. Our model is pre-trained from scratch for 600K iterations. We set the batch size to 8. 

\subsection{Comparison of Zero-Shot Generalization}

To demonstrate the improvement in model generalization, we conducted a cross-domain zero-shot experiment. We selected several state-of-the-art methods for comparison, including PSMNet~\cite{chang2018pyramid}, CFNet~\cite{shen2021cfnet}, GwcNet~\cite{guo2019group}, COEX~\cite{bangunharcana2021correlate}, FADNet++~\cite{wang2021fadnet++}, CasStereo~\cite{Gu_2020_CVPR_CasStereo}, IGEV~\cite{xu2023iterative}, Selective-IGEV~\cite{wang2024selective}, and StereoBase~\cite{guo2023openstereo}. All these methods were evaluated using their best-performing checkpoint, trained on the SceneFlow dataset.

\noindent\textbf{Quantitative Results.} As shown in Table \ref{tab:finetune-test}, we evaluates the effectiveness of our BooSTer. The training pipeline combines simulation data with real-world synthesized data to improve the generalization capabilities of the model. The results demonstrate a substantial improvement in the zero-shot performance, as evidenced by the significant increase in the model's accuracy and robustness across various test sets. Specifically, our method outperforms the baseline model, achieving a notable reduction in disparity estimation errors and providing more precise depth predictions. These results indicate that the mixed training approach not only improves the generalization of the algorithm but also enhances its performance in real-world scenarios, thereby validating the effectiveness of our approach in narrowing the domain gap and improving model robustness.

\noindent\textbf{Qualitative Results.} As shown in Figure \ref{fig:vis-dataset}, we present the qualitative results of stereo predictions from models trained on different datasets. In the first two rows, we demonstrate the impact of noise, where the disparity maps generated by StereoBase exhibit noticeable artifacts and inaccuracies, particularly in noisy regions. By incorporating a mixed data training pipeline, our method mitigates the effect of noise, improving disparity estimation and yielding clearer object boundaries. In the next two rows, we highlight the challenges of low-texture regions. StereoBase shows noticeable artifacts around textureless areas, such as poles and signs, while our method progressively refines depth accuracy, especially in regions lacking distinct texture features. In the final two rows, we focus on transparent or reflective materials, where StereoBase produces inaccurate depth maps due to the difficulty in handling such materials. Our method, however, enhances the results, providing more precise depth estimations even in the presence of complex surface reflections and transparency.

\begin{table*}[t]
\centering
\small
\begin{tabular}{llcccccccccc}
\toprule
\multirow{2}{*}{Networks} & \multirow{2}{*}{Training Dataset} & \multicolumn{3}{c}{ETH3D} & \multicolumn{3}{c}{KITTI12} & \multicolumn{3}{c}{KITTI15} \\
\cmidrule(lr){3-5} \cmidrule(lr){6-8} \cmidrule(lr){9-11} 
& & EPE $\downarrow$ & D1 $\downarrow$ & \textgreater2px $\downarrow$ & EPE $\downarrow$ & D1 $\downarrow$ & \textgreater2px $\downarrow$ & EPE $\downarrow$ & D1 $\downarrow$ & \textgreater2px $\downarrow$ \\
\midrule
\multirow{3}{*}{PSMNet} & SceneFlow & 8.272 & 8.799 & 10.74 & 4.683 & 30.512 & 43.35 & 5.986 & 32.149 & 44.47  \\
& MFS Dataset & 0.534 & 2.204 & 3.506 & 1.009 & 4.322 & 6.984 & 1.605 & 4.729 & 8.537  \\
\rowcolor{graycolor} \cellcolor{white}&  DiffMFS (Ours) & \textbf{0.397} & \textbf{1.531} & \textbf{2.481} & \textbf{0.859} & \textbf{4.077} & \textbf{6.667} & \textbf{1.052} & \textbf{4.507} &\textbf{8.312}  \\
\midrule
\multirow{3}{*}{CFNet} & SceneFlow  & \textbf{0.413} & 1.832 & 2.629 & 1.010 & 4.742 &  7.789 & 1.839 &  5.758 &9.939 \\
& MFS Dataset & 0.712 & 3.194 & 4.780 & 0.922 & 4.523 & 7.200 & 1.113 & 5.127 & 9.079  \\
\rowcolor{graycolor} \cellcolor{white}&  DiffMFS (Ours) & 0.443 & \textbf{1.434} & \textbf{2.500} & \textbf{0.881} & \textbf{4.169} &  \textbf{6.669} & \textbf{1.030} &  \textbf{4.711} & \textbf{8.562} \\
\midrule
\multirow{3}{*}{IGEV} & SceneFlow & 0.288 & 3.610 & 1.669 & 1.027 & 5.135 & 7.714 & 1.212 & 6.034 & 9.653 \\
& MFS Dataset & 0.703 & 4.657 & 2.105 & 0.824 & 3.546 &5.530 & 1.061 & 4.912 & 8.625 \\
\rowcolor{graycolor} \cellcolor{white}& DiffMFS (Ours) & \textbf{0.284} & \textbf{3.124} & \textbf{1.164} & \textbf{0.803} & \textbf{3.505} & \textbf{5.438} & \textbf{1.001} & \textbf{4.659} & \textbf{8.149}  \\
\midrule
\multirow{3}{*}{StereoBase} & SceneFlow & 0.286 & 1.284 & 1.762 & 1.022 & 4.983 & 8.091 & 1.199 & 5.477 & 9.478 \\
& MFS Dataset & 0.363 & 1.731 & 2.343 & 0.867 & 4.008 & 6.571 & 1.409 & 4.273 & 7.874 \\
\rowcolor{graycolor} \cellcolor{white}& DiffMFS (Ours) & \textbf{0.250} & \textbf{1.047} & \textbf{1.392} & \textbf{0.777} & \textbf{3.468} & \textbf{5.475} & \textbf{0.984} & \textbf{3.904} & \textbf{7.176}  \\
\bottomrule
\end{tabular}
\caption{Zero-shot validation results of models pre-trained on our generated dataset compared with synthetic dataset and data generation method. The results indicates that our stereo data generation method outperforms the state-of-the-art datasets. }
\label{tab:zero-shot}
\end{table*}



\subsection{Comparison of Data Generation Methods }

To evaluate the effectiveness of our stereo image generation method, we begin by comparing it to the synthetic dataset SceneFlow~\cite{mayer2016large} and the real-world image-based dataset generation method MFS~\cite{watson2020learning}. Four distinct model architectures are trained on each dataset using the same training settings.

As shown in Table \ref{tab:zero-shot}, our method improves generalization across multiple datasets and network architectures, with notable performance gains on the outdoor KITTI datasets. These improvements stem from the enhanced inpainting module, which ensures more coherent content in the generated stereo images, leading to better depth estimation, especially in occlusions and boundary transitions.

\subsection{Ablation Study}
In this section, we conduct a series of ablation studies to analyze the impact of different module choices in our method.

\noindent\textbf{Dataset Mixture.} We evaluate the effect of different data mixing strategies on the zero-shot performance, testing three strategies: synthetic data, simulation data combined with single-view images, and a mix of simulation, single-view, and real-world images. The results show that stereo images generated from single-view images notably boosts zero-shot performance, suggesting it provides data closer to real-world conditions, helping to bridge the domain gap. Although including real-world data further improves performance, the gain is modest, indicating that its impact is incremental compared to the substantial improvement seen with generated stereo data. This is likely due to the smaller scale and more specific nature of the real-world data used.

\noindent\textbf{Inpainting method.} When applying forward warping method, holes in the generated image indicate regions where no pixel is matched to the single-view image. As shown in Table \ref{tab:ablations}, using an advanced inpainting model yields better performance compared to the random hole-filling strategy in~\cite{watson2020learning}. However, Stable Diffusion often creates confusion between the object and background edges, reducing the image quality. In contrast, our Edge-Aware Filling module produces more visually coherent content, resulting in smoother object-background transitions and improved depth estimation accuracy. More visualization are shown in the supplementary materials.

\noindent\textbf{Stereo Matching Model.} To evaluate the performance of the proposed hybrid feature encoder, we applied the same training strategy to the state-of-the-art network, StereoBase~\cite{guo2023openstereo}. When compared to our model, which was trained under identical conditions, our approach demonstrates a effective improvement in zero-shot generalization performance, thereby validating the effectiveness of the encoder optimization proposed in this work.

\begin{table}[t]
\centering
\small
\begin{tabular}{llcc}
\toprule
\multirow{2}{*}{Exps} & \multirow{2}{*}{Methods} & \multicolumn{2}{c}{KITTI15} \\
\cmidrule(lr){3-4}
& & EPE $\downarrow$ & D1 $\downarrow$ \\
\midrule
\multirow{4}{*}{Data} & SF & 1.12 & 5.03\\
& MONO & 0.92 & 3.78 \\
& SF+MONO & 0.90 & 3.37 \\
& \underline{SF+MONO+DS} & \textbf{0.89} & \textbf{3.22}\\
\midrule
\multirow{3}{*}{Inpainting} & No & 0.99 & 3.98\\
& Stable-Diffsuion &  0.93 & 3.56\\
& \underline{Ours} & \textbf{0.89} & \textbf{3.22}\\
\midrule
\multirow{2}{*}{Model}
& StereoBase & 0.91 & 3.45 \\
& \underline{Ours} &\textbf{0.89} & \textbf{3.22}  \\
\bottomrule
\end{tabular}
\caption{Ablation experiments. Settings used in our method are \protect\underline{underlined}. ``SF" indicates Sceneflow, ``MONO" indicates real-world single-view images and ``DS" indicates DrivingStereo.}
\label{tab:ablations}
\end{table}

\section{Conclusion}


In this paper, we propose a novel framework for stereo matching, \textbf{BooSTer}, which addresses the challenges of limited labeled data and domain gaps between synthetic and real-world images. By leveraging large-scale mixed image sources, including synthetic, real, and single-view datasets, our approach significantly improves model generalization and performance. We introduce a data generation strategy that combines DFM and diffusion models, enabling the generation of dense stereo matching data from single-view images. Additionally, to overcome the sparse labeled data in real-world datasets, we employ pseudo-mono depth labels and a dynamic scale- and shift-invariant loss function for enhanced supervision. Furthermore, we propose a hybrid encoder structure that integrates VFM, enabling direct knowledge transfer from VFMs at feature level. This approach improves the diversity and generalization of the encoder's features. Through these novel designs,our approach achieves superior performance on real-world zero-shot experiment, scenting an advancement in stereo matching.

\bibliographystyle{named}
\bibliography{ijcai25}

\end{document}